# Evolutionary computational platform for the automatic discovery of nanocarriers for cancer treatment

Namid Stillman*[1], Igor Balaz*[2], Antisthenis Tsompanas[3], Marina Kovacevic[4], Sepinoud Azimi[5], Sebastien Lafond[5], Andrew Adamatzky[3], Sabine Hauert[1]

[1] Department of Engineering Mathematics, University of Bristol, Bristol, UK
[2] Laboratory for Meteorology, Physics, and Biophysics, Faculty of Agriculture, University of Novi Sad, Novi Sad, Serbia
[3] Unconventional Computing Laboratory, University of the West of England, UK
[4] Department of Chemistry, Biochemistry, and Environmental Protection, Faculty of Sciences, University of Novi Sad, Novi Sad, Serbia
[5] Faculty of Science and Engineering, Åbo Akademi University Turku, Finland

*These authors contributed equally to the paper.

## Abstract

We present the EVONANO platform for the *evo*lution of *nano*medicines with application to anti-cancer treatments. EVONANO includes a simulator to grow tumours, extract representative scenarios, and then simulate nanoparticle transport through these scenarios to predict nanoparticle distribution. The nanoparticle designs are optimised using machine learning to efficiently find the most effective anti-cancer treatments. We demonstrate our platform with two examples optimising the properties of nanoparticles and treatment to selectively kill cancer cells over a range of tumour environments.

**Keywords:** in silico medicine; drug discovery; nanomedicine; multi-scale modelling; machine learning; evolutionary optimisation

1. Introduction

Cancer is known to be a complex and multiscale disease, where tumour growth is caused by multifactorial effects ranging from individual cell stressors, mutations in cell signalling pathways, changes in the local tumour microenvironment, and the overall disruption of tissue homeostasis (Basanta & Anderson, 2017; Deisboeck et al., 2011; Sawyers, 2004; Sever & Brugge, 2015). Nanoparticle-based drug vectors have the potential for improved targeting of cancer cells when compared to free drug delivery through the design of cell-specific binding moieties and the encapsulation of drugs within nanoparticles that improve bio-transport (Bazak et al., 2015;

Roberts & Palade, 1995). Furthermore, nanoparticles can be used to create novel treatment strategies that rely on, for example, delayed release, local photo-thermal therapies, and treatments that take advantage of the collective dynamics of the nanoparticles (Hauert & Bhatia, 2014; Tong et al., 2012; von Maltzahn et al., 2011).

Novel anti-cancer nanomedicines are possible due to the expansive range of design parameters that can be altered. These include the four 'S's (size, shape, stiffness, and surface coating) as well as more complicated designs, such as encapsulated nanoparticles, self-assembling nanoparticles, and even nanoparticles that are capable of local computation (Fu et al., 2018; Kumari et al., 2016; Li et al., 2016; Simberg et al., 2007). However, given the vast parameter space, effective design of nanoparticles is a considerable challenge, as each of the chosen particle parameters impacts how they behave in the body (Hauert & Bhatia, 2014; Blanco et al., 2015; Angioletti-Uberti, 2017; Stillman et al., 2020). To improve nanoparticle design, *in silico* models have been shown to be efficient at both the prototyping and hypothesis testing stage, preventing costly trial-and-error search routines for potential candidates to test (Ginsburg & McCarthy, 2001; Goldie, 2001; Hauert et al., 2013). However, these *in silico* models often focus on singular aspects of the nanoparticles' journey through the body and are unable to systematically generate virtual tumour scenarios.

In this paper, we present a new platform for drug discovery, EVONANO, that combines models at the scale of an individual cell to the growth dynamics of a virtual tumour, whilst also applying machine learning to more efficiently explore the nanoparticle design space. The tumour growth dynamics are simulated using PhysiCell, an open-source package for simulating large cellular systems (Ghaffarizadeh et al., 2018) . These simulations create a virtual tumour using an agent-based model (ABM) which allows for high level of adaptability, such as the inclusion of heterogeneity in the cell phenotypes (cancer cells, cancer stem cells, macrophages). The virtual tumour models are then used to generate representative scenarios for trialling nanoparticles *in silico*.

Modelling nanoparticles over a whole tissue is computationally expensive and limits the degree in which nanoparticle parameters influence tumour dynamics. Instead, we systematically designate representative sections of tissue which are then used in a stochastic simulator, the STochastic Engine for Pathways Simulations (STEPS), which allows for simulations of stochastic networks over complex spatial domains through the discretisation of the well-mixed domain into smaller well-mixed subunits (known as voxels) (Chen & de Schutter, 2017; Hepburn et al., 2009, 2012). We specifically focus on a stochastic simulator to capture the inherent randomness that occurs within biochemical reaction networks. Nanoparticle-cell interactions are modelled using such reaction networks, where complicated signalling pathways can be introduced using the systems biology markup language (SBML) standard (Hucka et al., 2003). Finally, optimisation of nanoparticle design is done using a custom-built evolutionary algorithm (Preen et al., 2019; Tsompanas et al., 2020a, Balaz et al., 2020).

We present two examples demonstrating the EVONANO platform. The first designs nanoparticles that are able to penetrate an optimum distance to cover the majority of what is assumed to be a homogenous tumour mass. We optimise over the concentration, size, binding affinity, and payload of nanoparticles. We show how an evolutionary algorithm is able to effectively choose parameters that result in more than 90% of cancer cells successfully treated with low overall injected dosage. We then give an example of optimisation of treatments targeting a heterogeneous tumour mass, containing both cancer cells and cancer stem cells and where cancer stem cells are a minor cell population that are responsible for treatment resistance, metastatic growth and tumour recurrence, providing a promising target for nanomedicine and anti-cancer therapies (Gener et al., 2016). We increase the complexity of the design space by introducing two nanoparticle drug vectors, one that is specifically lethal to cancer cells and a second that is lethal to cancer stem cells only. In both of these examples, we find parameters that can preferentially kill cancer cells (and cancer stem cells) whilst minimising the overall dosage.

We end with a discussion of current and future developments for the EVONANO platform, such as integrating molecular dynamic simulations and validation using *in vitro/vivo* experiment and clinical or patient data, before making our concluding remarks.

2. Material and methods

The EVONANO platform uses modular design principles to implement and explore multiscale simulations over large parameter spaces[1]. The pipeline consists of three central modules: simulation of a virtual tumour, simulation of nanoparticle-tissue interactions, and evolutionary optimisation routines for nanoparticle design. Figure 1 illustrates an abstract workflow of the EVONANO platform.

---

[1] Code is available at https://bitbucket.org/hauertlab/evonano_methods/src/master/

*2.1 Virtual Tumour Module*

We begin by generating a virtual tumour, as shown in Figure 2. The virtual tumour models a representative tumour grown under certain assumptions. For example, here we focus on two aspects of tumour growth, the vasculature of the tumour and the inclusion of cancer stem cells. Tumour-specific features, such as the initial size and distribution of cells, structure of the vasculature, and existence of resistant sub-populations, can all be modelled in the virtual tumour. The virtual tumour is then used to generate relevant biological scenarios to test nanoparticles on. Here we consider the penetration distance of nanoparticles into tissue.

We used a modified version of the open-source PhysiCell platform, a cross-platform agent-based modelling framework for 2-D and 3-D multicellular simulations (Ghaffarizadeh et al., 2018). Agent-based models allow for high adaptability by modelling the interactions of individual sub-units (cells) which follow some system of rules and relevant dynamics. PhysiCell combines two modelling approaches: agent-based at the individual cell level and reaction-diffusion calculations for modelling diffusing substrates (Ghaffarizadeh et al., 2016). Each cancer cell is represented as individual agent and agents are associated with a library of sub-models for simulating cell, fluid, and solid volume changes, cycle progression, apoptosis, necrosis, mechanics, and motility. Since PhysiCell only supports differentiated cancer cells (CCs), we also modified its source code to include cancer stem cells (CSCs) and vasculature growth, modelled as vessel points (VPs).

We generate two virtual tumours, one with only cancer cells (homogenous) and one with both cancer cells and cancer stem cells (heterogeneous). The virtual tumour is initialised in PhysiCell with a single cancer cell (CC). We assume that the tissue surrounding the cancer cells is made up of healthy cells and that the space between cells within the tumour represents the tumour extracellular matrix (ECM). The virtual tumour is grown until it reaches the size of approximately 500,000 CCs at which point the simulation is stopped and we use simulation output to generate scenarios for the nanoparticle-tissue simulations.

The CSCs have an altered cellular behaviour so that, in contrast to CC, they cannot enter apoptotic or necrotic state. When in conditions of reduced oxygen concentration, they became dormant. CSC are generated either by dedifferentiation of CC or by (symmetric or asymmetric) division of CSC. In the case of dedifferentiation two daughter cells are created: 1 CC and 1 CSC. There are no reliable empirical data that describe the relative ratio of normal CC division to dedifferentiation, so we used a fixed probability of 99.5% (normal CC division) and 0.5% (dedifferentiation). CSC division can either be asymmetric giving a generation of 1 CC and 1 CSC, or symmetric and generating either 2 CSCs or 2 CCs. We set the probability of asymmetric division of CSC to 99%. The symmetric division, with probability of 1%, is further divided into 99% probability of creating 2 CSCs, and 1% of creating 2 CCs. We find that this distribution of probabilities leads to a virtual tumour with approximately 1% of CSC, which is consistent with experimental data (Bao et al. 2013).

To model the vascular network, we rely on a discrete approach which allows us to produce vascular networks of desired topologies. While it is possible to include a higher fidelity model of the interaction between nanoparticles and vascular flow such as a combination of CFD and agent-based or Brownian particles modelling, we chose an option that will minimally increase complexity of the pipeline. Therefore, we implemented a simplified model of vasculature directly into the PhysiCell by creating a new type of dividing agents with the following properties: they secrete oxygen; they are not movable and cannot enter apoptosis and necrosis; they are connected to each other apically. When dividing, vasculature agents can keep or change spatial direction by inserting a branching point. As a result, the growing vasculature is network-like. At the beginning of the simulation the following parameters are set: the initial position of vasculature agents, the maximum number of vasculature agents, and the frequency of branching.

We assume that nanoparticles enter into the tumour only from VPs, either through extravasation or trans-endothelial transport (Sindhwani, S. et al, 2020). Furthermore, we assume that the extravasation rate is constant across all VPs. We discuss extensions regarding the rate of extravasation in Section 4.

PhysiCell also models diffusing substrates using a multi-substrate diffusion solver, BioFVM, that divides the simulation domain into a collection of non-intersecting voxels. BioFVM supports diffusion, decay, cell-based secretions/uptake, and bulk supply/uptake functions for each individual substrate and is used here to capture the diffusion of oxygen and nutrients across the virtual tumour.

*2.2 Virtual Tissue Module*

To effectively capture the movement of nanoparticles through tissue, and their interaction with cells, stochastic simulations are used. We use the open-source package STEPS for simulating the cell-nanoparticle interactions. An example schematic of tissue module is given in Figure 3.

In this work, we restrict our focus to well-mixed compartments within domains that are the size of a single cell (compartment length L = 10um) and ignore the influence of spatiality as introduced by the discretisation of the domain into sub-volume elements (voxels). We model interactions between nanoparticles and cell using a stochastic Michaelis-Menten reaction network (Hauert et al., 2013; Peppas, 1996),

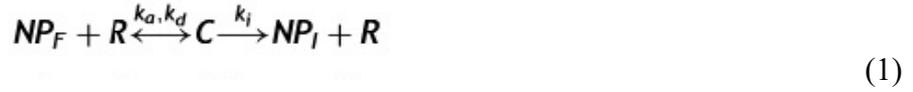

$$NP_F + R \xrightleftharpoons[k_d]{k_a} C \xrightarrow{k_i} NP_I + R \qquad (1)$$

where $NP_F$ is the free nanoparticles, R is the receptors on the cell membrane, C is nanoparticle-receptor complexes, and $NP_i$ is internalised nanoparticles. We assume that nanoparticles are able to actively target and bind to the receptors on CCs with binding rate, $k_a$, to release from receptors with dissociation rate $k_d$, and to internalise according to internalisation rate $k_i$. Nanoparticles move from neighbouring well-mixed compartments with probability $D/L^2$ where D is the diffusion coefficient of the nanoparticles and L is the compartment length. We also assume that nanoparticles have uniform binding to cell receptors and will explore alternate nanoparticle designs in future work.

To calculate the number of nanoparticles that are able to extravasate into the tumour within an idealised murine model, *NP0*, we use

$$NP0 = ID \, W \frac{PIDS^2L}{V_t} \frac{N_A}{ME}, \qquad (2)$$

where *ID* is the total injected dose of nanomedicine, *W* is the weight of the mouse, *PID* is the percentage of the injected dose that reaches the tumour, *S* is the characteristic length scale of the cell, *L* is the total required penetration depth, $V_t$ is the volume of the tumour, *M* the molecular mass of the nanoparticle payload, *E* the total number of payload molecules, and $N_A$ is Avogadro's constant. Parameter values are given in Table 1.

We assume that the cytotoxic effect of the nanoparticle is a function on the total payload carried by the nanoparticle, such that the threshold of nanoparticles required to induce cell death is given by

$$NP_{max} = \frac{PS^3 N_A}{E}, \qquad (3)$$

where P is the IC90 of the anti-cancer drug and we assume that the properties of the drug are not altered by the nanoparticle vector. Hence, altering E changes both the total dosage as well as the threshold of nanoparticles required to kill a cell.

For all simulations, we assume that nanoparticles circulate for 48 hours, that there is a constant release of nanoparticle from vessel points and that 1% of the injected dose will eventually extravasate into the tumour. These assumptions (circulation time, release rate and percentage injected dose that reaches the tumour) are based on previously published results on the modelling of nanoparticle dynamics (Hauert et al., 2013), and serve as a baseline to demonstrate the platform. Future work will seek to validate them using experiment data. When testing nanoparticle designs, we adjust both the number of payload molecules per nanoparticle (E), as well as the total number of nanoparticles within the treatment dose (NP0). Altering both of these reflects changes to the total injected dose that is administered to the mouse model. We aim to design nanoparticles that can penetrate deep into the tumour tissue but where the injected dose is low.

*2.3 Machine Learning Module*

Nanoparticles are highly customisable, covering a large design space of properties such as their size, material, shape, stiffness, binding moieties and charge (Hauert & Bhatia, 2014). Simulations at the tissue scale allow for the simulation of multiple particles as well as cell types. Nanoparticles can be optimised with different binding properties and diffusion coefficients, or drugs and targeting moieties specialised for individual cell types. For example, nanoparticles can be introduced that have specific binding to cancer cells but with higher internalisation rate for cancer stem cells.

This creates an extremely large parameter space to search in order to find the most effective treatment. The computational cost of simulation is prohibitive of unbound multiple tests and intelligent optimisation tactics are required instead. Consequently, the optimisation of the treatment parameters can be implemented with an amalgam of conventional, well-established algorithms, such as genetic algorithms (Preen et al., 2019), and unconventional, innovative methods (such as novelty search (Tsompanas et al., 2020a, Balaz et al., 2020) and metameric representations (Tsompanas et al., 2020b)).

As a first instance, we used an evolutionary algorithm which operates as follows: A population of randomly initialised individuals is produced, where each individual represents one solution (group of parameters) defining the functionality of the nanoparticles. These individuals are evaluated using the tissue module described in Section 2.2, with an appropriate score assigned to each individual based on the proportion of cell types killed and the overall drug dosage. The score represents the fitness of the individual and is described in more detail below. Given the assigned fitness, the best individuals are then selected through a tournament procedure and the crossover operator is applied to produce offspring individuals. These offspring individuals are then updated with the mutation operator and represent the next generation of individuals. This population is then evaluated again to find a new fitness and the process repeated until a predetermined number of generations has passed.

An example schematic of the evolutionary algorithm is shown in Figure 4. The parameters of the algorithm are as follows; the search space is initially four-dimensional (having four parameters to optimise), the population size is P=20, the tournament size for replacement and selection procedures is T=2, while the mutation operator is applied with probability of p=0.2 and it alters one parameter with random step size of s= $[-5; 5]$ %. Finally, the evolution of the population of individuals lasts for 100 generations.

*2.4 Scenarios for automatic nanocarrier discovery*

We next describe how the EVONANO platform is used to develop nanoparticle designs for penetrating tumour tissue in realistic biological scenarios. We consider nanoparticles within the range of binding strengths and diffusion coefficients previously investigated in (Hauert et al., 2013), and consider two examples: where the tumour contains a single cell type (homogenous) and where it contains two distinct sub-population of cell types (heterogeneous).

In both examples, we simulate a virtual tumour, using methods described in Section 2.1, to generate biologically relevant tumour scenarios in which the drug-loaded nanoparticles are required to successfully overcome. As in Hauert et al. (2013), we are primarily focused on maximising penetration of nanoparticles away from a vessel point. To develop realistic scenarios for testing nanoparticle design, we create tissue profiles using the virtual tumour, as shown in Figure 2 (b-c).

We first calculate the minimum distance to VPs for all CCs and use this to find the distance which would cover account for 95% of CCs being reached, given maximum penetration. We find this to be approximately 220μm, or the distance of 22 cells, for both homogenous and heterogeneous tumour scenario, shown in Figure 2 (d). Similar penetrations depths have been highlighted in experimental work by Jain et al. (2010).

Having calculated the necessary penetration distance required for maximal nanoparticle coverage, we then create simple scenarios consisting of 22 cells, generated from the PhysiCell simulations. We first randomly choose a VP from which we assume nanoparticles will release. We measure a distance of 220μm from this VP, where we assume each cell is 10μm. This distance, which represents the necessary penetration distance that nanoparticles are required to cover, is split into compartments of cubic compartments of length 10μm. The cell type within this compartment reflects the PhysiCell cell definition, such as VP, CC, CSC. In this work, we specify regions without cells as extracellular matrix (ECM). These 22 cells, spanning 220μm, relate to a test scenario for a nanoparticle design and is passed to the tissue module of EVONANO. Examples of both cell sampling and resulting scenarios are given in Figure 2(e-f).

Using the process described above, we generate 100 representative tissue-scale scenarios which are used for optimising nanoparticle design. However, in some scenarios, such as where multiple VPs are present, nanoparticle designs can become sufficiently under-constrained such that solutions have a misleadingly high fitness. As a result, we also choose a single scenario that represents the 'worst-case' scenario. For the homogenous case, this is one VP and 22 CCs without any ECM or VPs, while for the heterogeneous case, we choose one VP with 20 CCs and 2 CSCs, where the first CSC is 140μm from the VP and the second CSC is 180μm from the VP. Both these were chosen by observing which scenarios consistently had lower fitness than other scenarios when randomly sampling. After finding an optimum solution using either random sampling or the 'worst-case' scenario, we then evaluated the solution on all 100 generated scenarios.

### 2.4.1 *Optimised Tissue Distribution*

We first consider the homogenous case, where the tumour contains only cancer cells (CCs) and where we aim to find nanoparticle designs that cover the largest volume of the tumour and kill the highest proportion of CCs while keeping dosage low.

To grade the success of nanoparticle designs, we calculate the maximum number of nanoparticles that a cell can internalise before cell death is triggered. This is a function of the number of drug molecules, E, carried by the nanoparticle, given in Eq. (3). To demonstrate our methods, we consider nanoparticles that are loaded with Doxorubicin, a well-known and relatively common anti-cancer drug. We assume an IC90 of Doxorubicin to be 10μM (Abraham, et al., 2004) and that the nanoparticles can carry between 100 to 10,000 drug molecules. This corresponds to a maximum lethal dose, per cell, of between 600 to 60,000 nanoparticles. We include large ranges to allow for the machine learning module to search through the state space in order to find an optimal solution.

Here, we look to optimise two nanoparticle -specific parameters; the diffusion coefficient (D) and the binding affinity ($k_a$), relating to the respective size and, for example, a nanoparticle targeting ligand that is agnostic to cell type. We consider the diffusion coefficient, D, to be in the range of $10^{-8}$ to $10^{-6}$ cm$^2$/s. Using the Stokes-Einstein equation (Lindsay, 2010) and assuming biologically realistic values for viscosity of the surrounding medium, this reflects 10nm to 1um sized nanoparticles. We vary the binding affinity, $k_a$, of $10^3$ to $10^6$ (Ms)$^{-1}$, while keeping disassociation rate, $k_d$, and internalisation rate, $k_i$, for particles fixed at $10^{-4}$ s$^{-1}$ and $10^{-5}$ s$^{-1}$, respectively. This relates to a disassociation constant, $K_D = k_d/k_a$, of between $10^{-1}$ to $10^2$ nM.

We also consider the influence of two treatment-specific parameters, the concentration of nanoparticles within the treatment dose as well as the number of molecules per nanoparticle. As described above, we assume that extravasation rates are equal across all VPs, such that scenarios with additional VPs will have a linear increase in the total number of nanoparticles released. The change in concentration in nanoparticles will result in a different number of nanoparticles released from the VPs. We explore nanoparticle concentration within the range of 10,000 to 1 million nanoparticles, which we assume circulates for 48 hours and that 1% of the nanoparticles reach the tumour, as discussed in Section 2.2. We also assume that nanoparticles can carry between 100 to 10,000 drug molecules, giving an upper and lower bound for dosage of 0.025 to 250 mg/kg, calculated using Eq. (2). However, dosages higher than 55 mg/kg of Doxorubicin have been reported to lead to toxic effects in mice (Parr et al, 1997). As a result, we do not evaluate any parameter solutions that result in dosage higher than 55 mg/kg.

Finally, given the test scenarios and parameters above, we use the machine learning module to optimise nanoparticle design. The fitness of each individual parameter set for each test scenario is calculated as the proportion of cancer cells killed minus the total dosage (normalised to between 0 and 1),

$$Fitness = w \frac{CC}{NCC} - (dosage\ constraint) \qquad (4)$$

where *CC* is the number of killed cancer cells and *NCC* is the total number of cancer cells before treatment. Hence, a higher fitness represents both more cancer cells killed and a lower overall dosage. The average fitness is then calculated across all 5 scenarios. We also test how important the proportion of cancer cells killed is by multiplying this term by a weighting factor, *w*, but find that this does not greatly impact results. We run the optimisation routine three times for each different weight and either by randomly selecting five scenarios and taking the average fitness across these or using the single 'worst-case' scenario, as described in Section 2.4.

### *2.4.2 Cancer Stem Cell Treatment*

In Section 2.4.1, we describe a homogenous tumour in which there is only one cancer cell type which has shared features across the tumour. These CCs have no difference in their differentiation and are assumed to be equally relevant targets for nano-carriers. However, other cell types are known to play an important role in the growth of solid tumours, including development of resistance and the promotion of metastasis, such as cancer stem cells (Gener et al., 2016). Hence, we use the extension to the PhysiCell codebase, described in Section 2.1, to generate a heterogeneous virtual tumour which contains both cancer cells and cancer stem cells (CSCs) and then optimise nanoparticles that are lethal to this cell type only.

To compare optimisation across the two virtual tumours populations, both homogenous and heterogeneous, we keep optimisation parameters the same. We are interested in altering the diffusion and binding coefficient of the nanoparticles as well as the initial concentration and drug payload for two nanoparticle types, where lethality of both nanoparticles depends on cell type. We assume that cell death of CCs is triggered only beyond a threshold of internalised NP1s (CC-specific nanoparticles), while cell death of CSCs occurs only beyond a threshold of internalised NP2s (CSC-specific nanoparticles). These CSC-specific nanoparticles are supported by work reported in, for example, (Gener et al., 2020). As before, we model both nanoparticle properties on Doxorubicin, such as the drug molecular mass and IC90 but note that other nanoparticle-drug models can be used.

We now optimise an 8-dimension parameter space, in which we alter the diffusion coefficient, binding affinity, nanoparticle concentration and number of drug molecules for each of the nanoparticles but where we do not assume cell-specific properties other than the number of nanoparticles required to induce cell death.

We use the same evolutionary algorithm as before, but now with adapted fitness function,

$$Fitness = w(\frac{CC}{NCC} + \frac{CSC}{NCSC}) - (dosage\ constraint) \quad (5)$$

where *CSC* is the number of killed cancer stem cells and NCSC is the total number of cancer stem cells before treatment within a scenario. We again run the optimisation routine three times, both taking the average fitness calculated across 5 randomly selected scenarios within one generation, as well as run only on the 'worst-case' scenario described in Section 2.4.

## 3. Results

We now present results from the two example scenarios, as described in Section 2.4.1 and 2.4.2. We first consider a homogenous tumour cell population, where all nanoparticles have equal efficacy on CCs. We then consider a heterogeneous tumour cell population containing both CCs and CSCs.

For the first scenario, we use the EVONANO platform to generate a virtual tumour, as described in Section 2.1. We use our sampling procedure, described in Section 2.4, to generate scenarios that are optimised using methods described in Section 2.3 and evaluated at the tissue scale, as described in Section 2.2. Using the EVONANO platform, we find parameter values that result in 99% of all CCs killed within 95% of VPs with a dosage of 7.8 mg/kg. Optimum nanoparticle-based treatment parameters are given in Table 2.

Overall, more effective solutions (which kill more CCs) were found when nanoparticle treatments were tested on the worst-case scenario representing one VP and 22 CCs. This is likely due to the presence of multiple VPs in some scenarios, increasing the number of nanoparticles in the system and leading to fewer constraints on nanoparticle parameters. As a result, we focus on only this scenario, as solutions tended to have low dosage (10mg/kg and below) whilst still killing a large number of CCs.

We show results from running the evolutionary optimisation in Figure 4.b. We find that we are able to correctly identify the combination of parameters that allow for almost all cancer cells to be killed with low dosage. These parameters are D=1x10$^{-6}$ cm$^2$/s, k$_a$ = 700,000 (Ms)$^{-1}$ (equivalent to a nanoparticle radius of R = 2.5nm and K$_D$= 70nM), NP0 = 60,000 and E = 5,000, which has an overall fitness of 0.95. The nanoparticles that killed the largest number of cells followed the same trend as that observed in (Hauert et al., 2013), in which those with high diffusion and disassociation constants were found to penetrate and kill cells furthest from the vessel point. However, far fewer particles were shown to be required.

We next use the EVONANO platform to design nanoparticles within a heterogeneous tumour mass. In this scenario, a virtual tumour is generated using PhysiCell which consists of 527,000 cells with 517,000 CCs, 5,000 VPs and 5,000 CSCs. The CSCs have altered cellular behaviour as described in Section 2.2. We repeat the same sampling process for the generated tumour mass, resulting in 100 scenarios. Again, we find that the penetration distance from VPs required to cover 95% of the tumour mass is approximately 22 cells deep, with parameter values given in Table 2.

We use the machine learning module to find a parameter set that successfully kills 99% of CCs and 82% of CSCs within all scenarios. For NP1, D = 9.8x10$^{-7}$ cm$^2$/s, k$_a$ = 217,000 (Ms)$^{-1}$ (equivalent to a radius of R = 2.5nm and K$_D$= 22nM), NP0 = 923,000 and E = 400, while for NP2, D = 6.4x10$^{-7}$ cm$^2$/s, k$_a$ = 117,000 (Ms)$^{-1}$ (equivalent to a radius of R = 4nm and K$_D$= 12nM), NP0 = 150,000 and E = 2,500. This reflects a total dosage of 9.4 mg/kg for NP1 and 9 mg/kg for NP2.

This solution was found when the weighting constant was set to 5 and where nanoparticles were tested only on the worst-case scenario (described in Section 2.4). Multiple other solutions were found that achieved similarly high treatment success but with higher doses.

For nanoparticles tested on several randomly selected scenarios, we found that the solution would often fail to kill CSCs due to their low prevalence within the total cell population. However, when we heavily bias the gain in fitness according to cell death, by setting *w* in Eq. (5) to 10, we find a second solution that has significantly different nanoparticle parameters and where the total dosage is high. These parameters are, for NP1, D = 9.7x10$^{-9}$ cm$^2$/s, k$_a$ = 8,300 (Ms)$^{-1}$ (equivalent to a nanoparticle radius of R = 264 nm and K$_D$= 0.8nM), NP0 = 236,000 and E = 7,600, while for NP2, D = 5.7x10$^{-7}$ cm$^2$/s, k$_a$ = 791,000 (Ms)$^{-1}$ (equivalent to a radius of R = 4.5nm and K$_D$= 80nM),

NP0 = 828,000 and E = 1,200. This reflects a total dosage of 46.4 mg/kg for NP1 and 25.9 mg/kg for NP2, relatively high for nanomedicine treatment. However, this solution deviates from previous solutions and demonstrates that by tuning the fitness and randomly sampling from generated scenarios, we are able to find new and novel nanoparticle designs.

## 4. Discussions

In this paper, we have introduced the EVONANO platform for the automatic optimisation of nanoparticle design parameters in tumour tissue. This modular simulation tool builds on two open-source simulation frameworks, PhysiCell and STEPS, extending them to model tumour growth, extract representative worst-case tumour tissue scenarios, and then model nanoparticle penetration within these tumour tissues. Machine learning is used to achieve optimisation across scales.

In the first example, we show that we can use machine learning to find nanoparticle designs that are able to distribute and kill large proportions of tumour mass, while keeping dosage low. Overall, we find that high diffusion values (small nanoparticle radius) and low binding affinity (high $K_D$) are most effective, and that low nanoparticle concentration (tens of thousands of nanoparticles) will suffice, provided their payload is in the thousands. This allows nanoparticles to both penetrate the tumour, and avoid binding site barriers, as already demonstrated in previous work (Hauert et al., 2013). Alternatively, concentrations of several million nanoparticles can also be effective, and require a payload only of some several hundred drug molecules. The idea that high concentration of nanoparticles may lead to more effective cancer treatments complements other work that emphasises the importance of increasing nanoparticle dosage (Ouyang et al., 2020).

For the second example, consisting of a heterogeneous tumour with cancer stem cells, we successfully find nanoparticle designs that have high efficacy with low dosage. Again, we find that the solutions tend towards high diffusion coefficient (representing small nanoparticles), low binding affinity (representing high disassociation constants), high nanoparticle concentration and low drug molecules. This indicates that successful nanoparticle treatments may be those with high concentration of nanoparticles and low drug loading, as discussed above.

We also find a novel solution for the heterogeneous tumour: relatively large nanoparticles with high binding affinity. The diffusion coefficient and disassociation constant for these nanoparticles are within the parameter ranges that were expected to be unsuitable for nanoparticle designs, based on previous work in (Hauert et al., 2013). Instead, this solution demonstrates that nanoparticles with low diffusion coefficient and high binding affinity are able to successfully kill a high proportion of CCs, provided the total number of nanoparticles that reach the tumour is high. This reaffirms recent conclusions that the success of nanoparticle treatments may be highly dose dependent (Ouyang et al., 2020). This strategy requires a total dosage of 46 mg/kg for NP1, near to the assumed upper limit of 55 mg/kg, we note that this is based on the assumption of systemic circulation and that only 1% will make it to the tumour. For Doxil, it has been reported that the percentage injected dose that is able to make it to the tumour can be as high as 3.5% (Man et al., 2018), which would reflect, using Eq. (2), a reduced dosage of only 13 mg/kg. These findings, along with the design principles described above, highlights that it is possible to obtain more flexibility in nanoparticle-design.

In this paper, we present the first extension to homogenous cancer models through the inclusion of cancer stem cells. However, both cancer de-differentiated and stem cells were assumed to be homogenous in both size and shape. This limitation can be easily relaxed, with heterogeneous cells of any size, shape or type included within the virtual tumour for input for the optimisation routine. For example, stroma cells such as fibroblasts or cancer cells with increased resistance can be included. The impact of nanoparticle treatment can then be returned to the virtual tumour to assess how treatment impacts tumour progression, based on targeting some or all of these cell types. By adapting and including further biological details into the virtual tumour and nanoparticle properties, the EVONANO platform has the potential to link to clinical data, such as histology samples, and to develop and evaluate potential treatments *before* choosing a specific course of action for a patient. This would be a significant step towards the long sought-after objective of personalised (patient-specific) medicine (Ryu et al., 2014; Shin et al., 2013).

We have presented a new open-source platform that builds on both PhysiCell and STEPS. One of the benefits of the modular design of the EVONANO platform is that these two powerful open source-packages are used. For example, STEPS is able to simulate stochastic pathways across spatially discrete domains. These features allow for simulating nanoparticle interactions with whole-cell models where nanoparticle binding can be modelled at the resolution of individual receptor types, and nanoparticle transport and internalisation can be a function of cell cycle or the distribution of nanoparticles within intracellular compartments, all of which have been shown to impact nanoparticle treatment efficacy (Behzadi et al., 2017; Deng et al., 2019; Kim et al., 2012; M. Liu et al., 2017). Alternatively, extensions to the PhysiCell codebase allow for increased complexity in tumour models. The vasculature network added into PhysiCell allows us to more precisely control the spatial effect of nanoparticle extravasation. This could lead to the better parameterisation of nanoparticle delivery to the tumour and enables simulation of the effect of drugs that disrupt vasculature network (such as the tumour necrosis factor) or those that block vasculature growth factors (for example, Bevacizumab, Cabozantinib, Pazopanib). These aspects will be investigated in our future work. Alternatively, nanoparticles can be introduced that, when binding together, lead to larger particles with lower diffusion coefficients. Both of these reflect realistic nanoparticle designs that have been demonstrated *in vitro* (Hoshyar et al., 2016; Sun et al., 2017). Finally, both PhysiCell and STEPS supports SBML models. This means that both cell-based models, nanoparticle signalling, and reaction networks can be developed, validated and shared between researchers to greater increase standardisation within the field of nanomedicine (Macklin et al., 2018).

By disconnecting the model of tissue scale dynamics from the virtual tumour (which includes the dynamics of hundreds of thousands, if not millions, of individual cells), we decrease computing time and allow for large parameter spaces to be explored at the physio-chemically relevant scale for nanoparticle design. Furthermore, the ability to model the stochastic interactions of populations of nanoparticles allows for the specific inclusion of effects that are outside the domain of deterministic models, such as self-aggregation, stochastic oscillations, and other noise-induced behaviours (Bell et al., 2014; Borkowska et al., 2020; Grzybowski & Huck, 2016; X. Liu et al., 2013). This gives a rich design space that can be explored to search for novel and emergent nanoparticle designs (Hauert & Bhatia, 2014; Tsompanas et al., 2020a).

In the future, we plan to use High Performance Computing (HPC) as a natural solution to meet computational needs of the EVONANO framework. This will facilate the development of personalised treatments, and will allow for the exploration of more computationally complex environments. Approaches considered may include employing clusters for the computations. In such a setup, an ensemble of fully independent computing systems is used that are linked together to form a distributed computer system which enables parallel computing to be undertaken with standard hardware and software. Another approach is to use parallel "Graphics Processing Units (GPU)", a card-based device, which offers increased scalability and processing performance compared to a CPU. This approach provides a more affordable and efficient solution for highly expensive computational challenges.

We aim to increase the level of parallelisation within each module of the pipeline to execute the simulation processes simultaneously. For example, most of the STEPS simulations can be performed independently without network communication, and PhysiCell-EMEWS (Ozik et al., 2018) could be integrated into the EVONANO workflow.

The richness that can be accessed through the inclusion of nanoparticle-scale models also allows for downstream integration of molecular dynamics simulations. These simulations can, for example, explore the influence of a drug's chemical properties on the NP coating, the dependency of binding moieties on diffusion due to particle self-aggregatation or the formation of protein coronas (Babakhani, 2019; Treuel et al., 2015; Villaverde & Baeza, 2019, Kovacevic et al., 2020). By including molecular dynamics, specific rules for nanoparticle synthesis can be found such as the number of binding ligands on the nanoparticle surface. Furthermore, these simulations give another scale that can be optimised, whereby machine intelligence applied at the molecular scale learns overall impacts on tumour progression, as mediated by the nanoparticle-cell interactions.

Finally, the EVONANO platform presented here, can be used to develop future design principles that, when combined with high-throughput testing, would give an integrated and calibrated tool for testing efficacy of proposed anti-cancer therapies. For example, by including data from *in vitro* tools for measuring the extravasation properties of nanoparticles into tumour tissue, the impact of the extracellular matrix on nanoparticle transport dynamics, or the interactions of nanoparticles with individual cells (Cunha-Matos et al., 2016, McCormick et al., 2020). Additionally, *in vivo* measurements can be used to calibrate models of the virtual tumour to further increase the biological relevance of generated testing scenarios.

## 5. Conclusions

We have presented EVONANO, a new software platform for the design and optimisation of anti-cancer nanomedicine. Our open-source platform combines multi-scale and modular simulations with machine learning techniques, testing the properties of both nanocarriers and treatments against biologically relevant tissue scenarios, generated from an in silico virtual tumour. We first demonstrate our platform by finding nanoparticle treatments that lead to the successful treatment of 95% of cancer cells within the virtual tumour while minimising dosage. We then introduce a second cell type, cancer stem cells, into the virtual tumour and find parameters for a combined treatment that successfully kills 99% of cancer cells and more than 80% of cancer stem cells within the tumour tissue. The platform addresses a significant clinical challenge in automatically designing nano-carriers with appropriate transport properties. Future work will concentrate on combining our simulator with details from cell biology, such as the impact of other cell types on resistance, molecular dynamic simulations, and the validation and calibration of in silico results with *in vitro* and *in vivo* data.

## Acknowledgements

This project has received funding from the European Union's Horizon 2020 research and innovation programme under grant agreement No 800983.

## Conflicts of interest

The authors declare no competing financial interests.

## References

Abraham, S. A., McKenzie, C., Masin, D., Ng, R., Harasym, T. O., Mayer, L. D., & Bally, M. B. (2004). In vitro and in vivo characterization of doxorubicin and vincristine coencapsulated within liposomes through use of transition metal ion complexation and pH gradient loading. Clinical cancer research, 10(2), 728-738.

Angioletti-Uberti, S. (2017). Theory, simulations and the design of functionalized nanoparticles for biomedical applications: A Soft Matter Perspective. In *npj Computational Materials*.

Babakhani, P. (2019). The impact of nanoparticle aggregation on their size exclusion during transport in porous media: One- and three-dimensional modelling investigations. *Scientific Reports*.

Bae, Y. H., & Park, K. (2011). Targeted drug delivery to tumors: Myths, reality and possibility. In *Journal of Controlled Release*

Bao, B., Ahmad, A., Azmi, A. S., Ali, S., & Sarkar, F. H. (2013). Overview of cancer stem cells (CSCs) and mechanisms of their regulation: implications for cancer therapy. *Current protocols in pharmacology*, *61*(1), 14-25.

Balaz, I., Petrić, T., Kovacevic, M., Tsompanas, M. A., & Stillman, N. (2020). Harnessing adaptive novelty for automated generation of cancer treatments. *Biosystems*, 104290.

Basanta, D., & Anderson, A. R. A. (2017). Homeostasis back and forth: An ecoevolutionary perspective of cancer. *Cold Spring Harbor Perspectives in Medicine*.

Bazak, R., Houri, M., el Achy, S., Kamel, S., & Refaat, T. (2015). Cancer active targeting by nanoparticles: a comprehensive review of literature. In *Journal of Cancer Research and Clinical Oncology*.

Behzadi, S., Serpooshan, V., Tao, W., Hamaly, M. A., Alkawareek, M. Y., Dreaden, E. C., Brown, D., Alkilany, A. M., Farokhzad, O. C., & Mahmoudi, M. (2017). Cellular uptake of nanoparticles: Journey inside the cell. In *Chemical Society Reviews*.

Bell, I. R., Ives, J. A., & Jonas, W. B. (2014). Nonlinear effects of nanoparticles: Biological variability from hormetic doses, small particle sizes, and dynamic adaptive interactions. *Dose-Response*.

Blanco, E., Shen, H., & Ferrari, M. (2015). Principles of nanoparticle design for overcoming biological barriers to drug delivery. In *Nature Biotechnology*.

Borkowska, M., Siek, M., Kolygina, D. v., Sobolev, Y. I., Lach, S., Kumar, S., Cho, Y. K., Kandere-Grzybowska, K., & Grzybowski, B. A. (2020). Targeted crystallization of mixed-charge nanoparticles in lysosomes induces selective death of cancer cells. *Nature Nanotechnology*.

Chen, W., & de Schutter, E. (2017). Parallel STEPS: Large Scale Stochastic Spatial Reaction-Diffusion Simulation with High Performance Computers. *Frontiers in Neuroinformatics*, *11*(February), 1–15.

Cheng, Q., Wei, T., Farbiak, L., Johnson, L. T., Dilliard, S. A., & Siegwart, D. J. (2020). Selective organ targeting (SORT) nanoparticles for tissue-specific mRNA delivery and CRISPR–Cas gene editing. *Nature Nanotechnology*.

Cunha-Matos, C. A., Millington, O. R., Wark, A. W., & Zagnoni, M. (2016). Real-time assessment of nanoparticle-mediated antigen delivery and cell response. *Lab on a Chip*.

Deisboeck, T. S., Wang, Z., MacKlin, P., & Cristini, V. (2011). Multiscale cancer modeling. *Annual Review of Biomedical Engineering*.

Deng, H., Dutta, P., & Liu, J. (2019). Stochastic modeling of nanoparticle internalization and expulsion through receptor-mediated transcytosis. *Nanoscale*.

Fu, Y., Feng, Q., Shen, Y., Chen, M., Xu, C., Cheng, Y., & Zhou, X. (2018). A feasible strategy for self-assembly of gold nanoparticles: Via dithiol-PEG for photothermal therapy of cancers. *RSC Advances*.

Gener, P., de Sousa Rafael, D. F., Fernández, Y., Ortega, J. S., Arango, D., Abasolo, I., Videira, M., & Schwartz, S. (2016). Cancer stem cells and personalized cancer nanomedicine. In *Nanomedicine*.

Gener, P., Montero, S., Xandri-Monje, H., Díaz-Riascos, Z. V., Rafael, D., Andrade, F., ... & Schwartz Jr, S. (2020). Zileuton™ loaded in polymer micelles effectively reduce breast cancer circulating tumor cells and intratumoral cancer stem cells. *Nanomedicine: Nanotechnology, Biology and Medicine*, *24*, 102106.

Ghaffarizadeh, A., Heiland, R., Friedman, S. H., Mumenthaler, S. M., & Macklin, P. (2018). PhysiCell: An open source physics-based cell simulator for 3-D multicellular systems. *PLoS Computational Biology*.

Ginsburg, G. S., & McCarthy, J. J. (2001). Personalized medicine: Revolutionizing drug discovery and patient care. In *Trends in Biotechnology*.

Goldie, J. H. (2001). Drug resistance in cancer: A perspective. In *Cancer and Metastasis Reviews*.

Grzybowski, B. A., & Huck, W. T. S. (2016). The nanotechnology of life-inspired systems. *Nature Nanotechnology*.

Hauert, S., Berman, S., Nagpal, R., & Bhatia, S. N. (2013). A computational framework for identifying design guidelines to increase the penetration of targeted nanoparticles into tumors. *Nano Today*, *8*(6), 566–576.

Hauert, S., & Bhatia, S. N. (2014). Mechanisms of cooperation in cancer nanomedicine: Towards systems nanotechnology. In *Trends in Biotechnology*.

Hepburn, I., Chen, W., Wils, S., & de Schutter, E. (2012). STEPS: Efficient simulation of stochastic reaction-diffusion models in realistic morphologies. *BMC Systems Biology*.

Hepburn, I., Wils, S., & de Schutter, E. (2009). STEPS: reaction-diffsion simulation in complex 3D geometries. *BMC Neuroscience*.

Hoshyar, N., Gray, S., Han, H., & Bao, G. (2016). The effect of nanoparticle size on in vivo pharmacokinetics and cellular interaction. In *Nanomedicine*.

Hucka, M., Finney, A., Sauro, H. M., Bolouri, H., Doyle, J. C., Kitano, H., Arkin, A. P., Bornstein, B. J., Bray, D., Cornish-Bowden, A., Cuellar, A. A., Dronov, S., Gilles, E. D., Ginkel, M., Gor, V., Goryanin, I. I., Hedley, W. J., Hodgman, T. C., Hofmeyr, J. H., … Wang, J. (2003). The systems biology markup language (SBML): A medium for representation and exchange of biochemical network models. *Bioinformatics*.

Jain, R. K., & Stylianopoulos, T. (2010). Delivering nanomedicine to solid tumors. Nature reviews Clinical oncology, 7(11), 653.

Kim, J. A., Aberg, C., Salvati, A., & Dawson, K. A. (2012). Role of cell cycle on the cellular uptake and dilution of nanoparticles in a cell population. *Nature Nanotechnology*.

Kovacevic, M., Balaz, I., Marson D., Laurini E., & Jovic B. (2020) Mixed-monolayer functionalized gold nanoparticles for cancer treatment: atomistic molecular dynamics simulations study. *BioSystems. (Accepted)*

Kumari, P., Ghosh, B., & Biswas, S. (2016). Nanocarriers for cancer-targeted drug delivery. In *Journal of Drug Targeting*. https://doi.org/10.3109/1061186X.2015.1051049

Li, Y., Lian, Y., Zhang, L. T., Aldousari, S. M., Hedia, H. S., Asiri, S. A., & Liu, W. K. (2016). Cell and nanoparticle transport in tumour microvasculature: The role of size, shape and surface functionality of nanoparticles. In *Interface Focus*.

Lindsay, S. (2010). *Introduction to nanoscience*. Oxford University Press.

Liu, M., Li, Q., Liang, L., Li, J., Wang, K., Li, J., Lv, M., Chen, N., Song, H., Lee, J., Shi, J., Wang, L., Lal, R., & Fan, C. (2017). Real-Time visualization of clustering and intracellular transport of gold nanoparticles by correlative imaging. *Nature Communications*.

Liu, X., Chen, Y., Li, H., Huang, N., Jin, Q., Ren, K., & Ji, J. (2013). Enhanced retention and cellular uptake of nanoparticles in tumors by controlling their aggregation behavior. *ACS Nano*.

Macklin, P., Friedman, S. H., & Project, M. (2018). Open-source tools and standardized data in cancer systems biology. *BioRxiv*, 244319.

Man, F., Lammers, T., & de Rosales, R. T. (2018). Imaging nanomedicine-based drug delivery: a review of clinical studies. *Molecular imaging and biology*, *20*(5), 683-695.

McCormick, S. C., Stillman, N., Hockley, M., Perriman, A., & Hauert, S. (2020) Measuring nanoparticle penetration through a tissue-mimetic chip. (*Under Review*)

Ouyang, B., Poon, W., Zhang, Y. N., Lin, Z. P., Kingston, B. R., Tavares, A. J., ... & Chan, W. C. (2020). The dose threshold for nanoparticle tumour delivery. *Nature materials*, *19*(12), 1362-1371.

Ozik, J., Collier, N., Wozniak, J.M., Macal, C., Cockrell, C., Friedman, S.H., Ghaffarizadeh, A., Heiland, R., An, G. and Macklin, P., 2018. High-throughput cancer hypothesis testing with an integrated PhysiCell-EMEWS workflow. BMC bioinformatics, 19(18), p.483.

Parr, M. J., Masin, D., Cullis, P. R., & Bally, M. B. (1997). Accumulation of liposomal lipid and encapsulated doxorubicin in murine Lewis lung carcinoma: the lack of beneficial effects by coating liposomes with poly (ethylene glycol). *Journal of Pharmacology and Experimental Therapeutics*, *280*(3), 1319-1327.

Peppas, N. A. (1996). Receptors: models for binding, trafficking, and signaling. *Journal of Controlled Release*.

Preen, R. J., Bull, L., & Adamatzky, A. (2019). Towards an evolvable cancer treatment simulator. *BioSystems*.

Roberts, W. G., & Palade, G. E. (1995). Increased microvascular permeability and endothelial fenestration induced by vascular endothelial growth factor. *Journal of Cell Science*.

Ryu, J. H., Lee, S., Son, S., Kim, S. H., Leary, J. F., Choi, K., & Kwon, I. C. (2014). Theranostic nanoparticles for future personalized medicine. In *Journal of Controlled Release*.

Sawyers, C. (2004). Targeted cancer therapy. In *Nature*. https://doi.org/10.1038/nature03095

Sever, R., & Brugge, J. S. (2015). Signal transduction in cancer chemoprevention. *Cold Spring Harb Perspect Med*.

Shin, S. J., Beech, J. R., & Kelly, K. A. (2013). Targeted nanoparticles in imaging: Paving the way for personalized medicine in the battle against cancer. In *Integrative Biology (United Kingdom)*.

Simberg, D., Duza, T., Park, J. H., Essler, M., Pilch, J., Zhang, L., Derfus, A. M., Yang, M., Hoffman, R. M., Bhatia, S., Sailor, M. J., & Ruoslahti, E. (2007). Biomimetic amplification of nanoparticle homing to tumors. *Proceedings of the National Academy of Sciences of the United States of America*.

Sindhwani, S., Syed, A. M., Ngai, J., Kingston, B. R., Maiorino, L., Rothschild, J., ... & Wu, J. L. (2020). The entry of nanoparticles into solid tumours. *Nature materials*, *19*(5), 566-575.

Song, W., Popp, L., Yang, J., Kumar, A., Gangoli, V. S., & Segatori, L. (2015). The autophagic response to polystyrene nanoparticles is mediated by transcription factor EB and depends on surface charge. *Journal of Nanobiotechnology*.

Stillman, N. R., Kovacevic, M., Balaz, I., & Hauert, S. (2020). In silico modelling of cancer nanomedicine, across scales and transport barriers. In *npj Computational Materials*.

Sun, J., Liu, Y., Ge, M., Zhou, G., Sun, W., Liu, D., Liang, X. J., & Zhang, J. (2017). A Distinct Endocytic Mechanism of Functionalized-Silica Nanoparticles in Breast Cancer Stem Cells. *Scientific Reports*.

Tong, R., Hemmati, H. D., Langer, R., & Kohane, D. S. (2012). Photoswitchable nanoparticles for triggered tissue penetration and drug delivery. *Journal of the American Chemical Society*.

Treuel, L., Docter, D., Maskos, M., & Stauber, R. H. (2015). Protein corona - from molecular adsorption to physiological complexity. *Beilstein Journal of Nanotechnology*.

Tsompanas, M. A., Bull, L., Adamatzky, A., & Balaz, I. (2020a). Novelty search employed into the development of cancer treatment simulations. *Informatics in Medicine Unlocked*.

Tsompanas, M. A., Bull, L., Adamatzky, A., & Balaz, I. (2020b). In silico optimization of cancer therapies with multiple types of nanoparticles applied at different times. *Computer Methods and Programs in Biomedicine*, 105886.

Villaverde, G., & Baeza, A. (2019). Targeting strategies for improving the efficacy of nanomedicine in oncology. In *Beilstein Journal of Nanotechnology*.

von Maltzahn, G., Park, J. H., Lin, K. Y., Singh, N., Schwöppe, C., Mesters, R., Berdel, W. E., Ruoslahti, E., Sailor, M. J., & Bhatia, S. N. (2011). Nanoparticles that communicate in vivo to amplify tumour targeting. *Nature Materials*.

Whitley, D. (1994). A genetic algorithm tutorial. *Statistics and computing*, *4*(2), 65-85.

# Tables

**Table 1:** Description and value of the assumed parameters for simulation

| Symbol | Description | Unit | Value |
|---|---|---|---|
| $PID$ | Percentage injected dose | - | 1% |
| $T$ | Time at which PID is measured | hours | 48 |
| $W$ | Weight of murine model | grams | 20 |
| $P$ | Potency of payload (IC90 of doxorobuicin) | μM | 10 |
| $M$ | Molar mass of payload (doxorubicin) | g/mol | 543.52 |
| $V_t$ | Volume of the tumour | mm$^3$ | 125 |
| $S$ | Characteristic length scale of cell | μm | 10 |
| $N_R$ | Number of receptors per cell | - | $10^5$ |

**Table 2:** Description and values of optimum nanoparticle treatment parameters. *The dosage is capped at 55 mg/kg to discount parameter values that lead to fatal treatments.

| Symbol | Description | Unit | Range | Homogenous Solution 1 NP1 | Heterogeneous Solution 1 NP1 | Heterogeneous Solution 1 NP2 | Heterogeneous Solution 2 NP1 | Heterogeneous Solution 2 NP2 |
|---|---|---|---|---|---|---|---|---|
| $D$ | Diffusion coefficient | cm2/s | [$10^{-8}$, $10^{-6}$] | 1x10-6 | 9.8x10$^{-7}$ | 6.4x10$^{-7}$ | 9.7x10$^{-9}$ | 5.7 x10$^{-7}$ |
| $k_a$ | Binding affinity | 1/Ms | [$10^3$, $10^6$] | 700,000 | 217,000 | 117,000 | 8,300 | 791,000 |
| $NP0$ | Nanoparticle concentration | molecules | [$10^4$, $10^6$] | 60,000 | 923,000 | 150,000 | 236,000 | 828,000 |
| $E$ | Number of drug molecules per nanoparticle | molecules | [$10^2$, $10^4$] | 5,000 | 400 | 2,500 | 7,600 | 1,200 |
| $ID$ | Dosage | mg/kg | [0.025, 55*] | 7.8 | 9.4 | 9.0 | 46.4 | 25.9 |
| $r$ | Nanoparticle radius | nm | | 2.5 | 2.5 | 4.0 | 264 | 4.5 |

# Figures

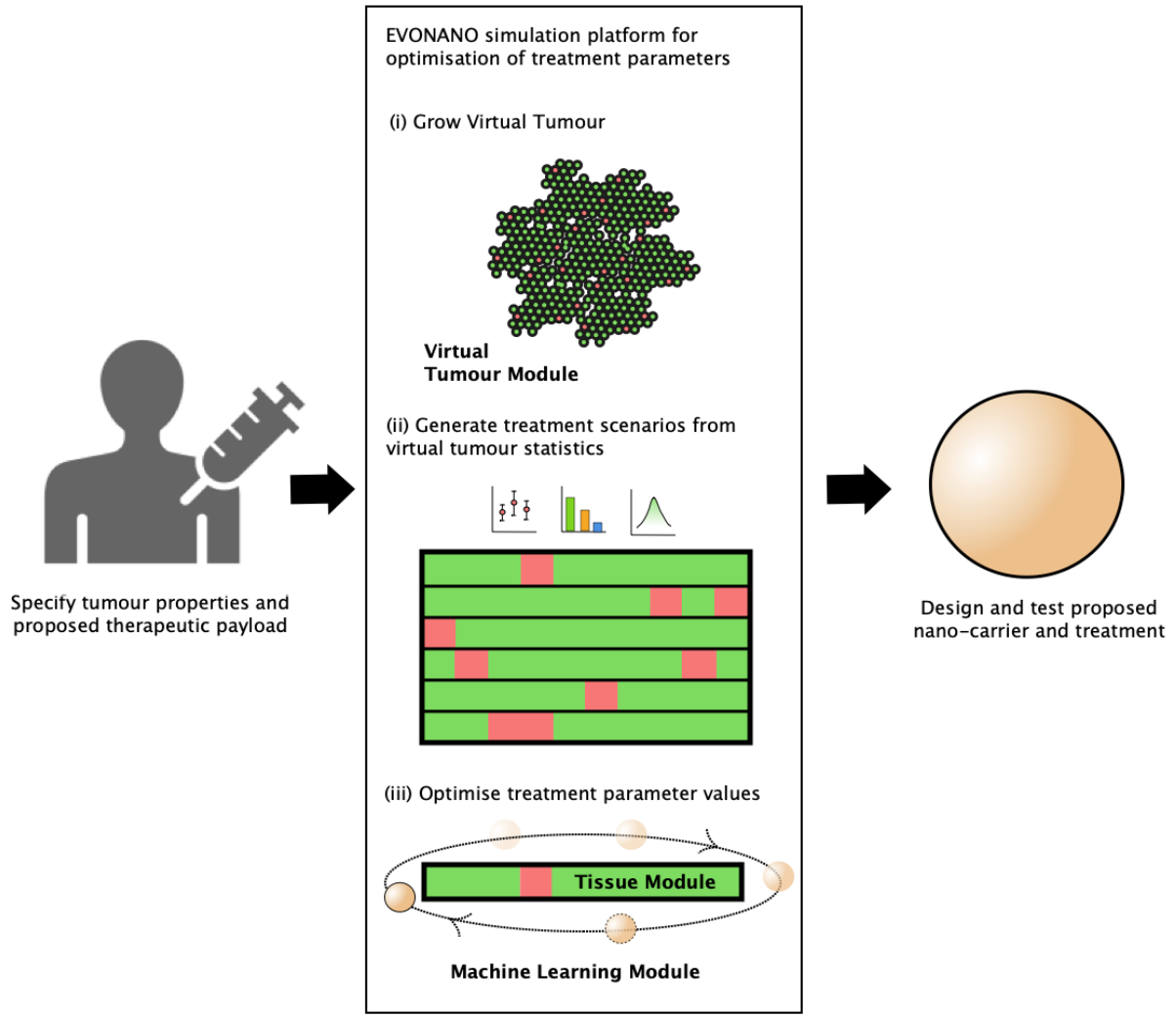

**Figure 1:** We present a general overview of the EVONANO platform. We begin by (A) specifying tumour and possible nanocarrier properties which are then used as assumptions within the EVONANO simulation platform. This proceeds as follows: (i) we first grow a virtual tumour to a sufficient size using the virtual tumour model (Section 2.1), (ii) summary statistics, such as necessary penetration distance, are calculated from the virtual tumour and used to generate representative treatment scenarios (Section 2.4), (iii) we then optimise the parameter values using the tissue module and machine learning module (Sections 2.2 and 2.3, respectively). The nano-carrier and treatment parameters can then be designed and tested using *in vitro/vivo* methods.

(a)

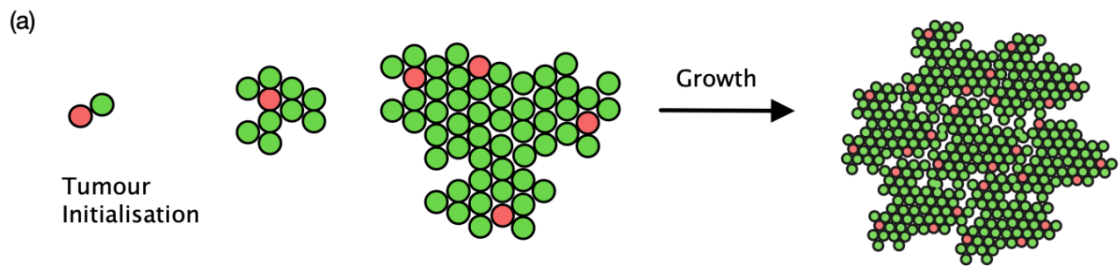

(b) **t = 1 day**

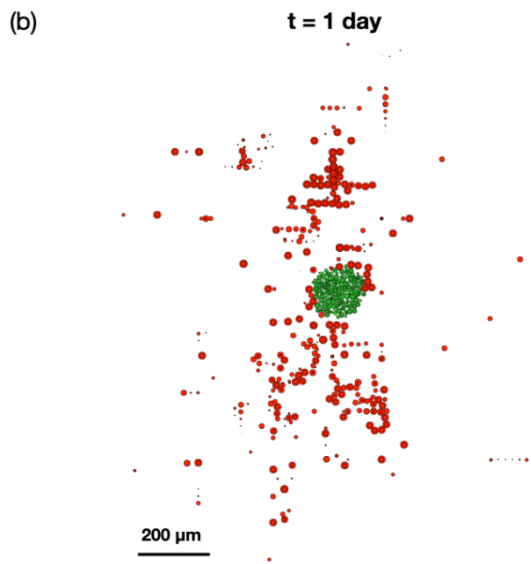

(c) **t = 7 days**

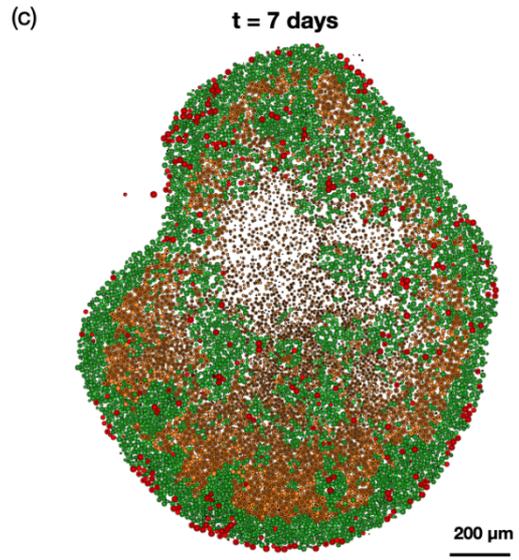

(d)

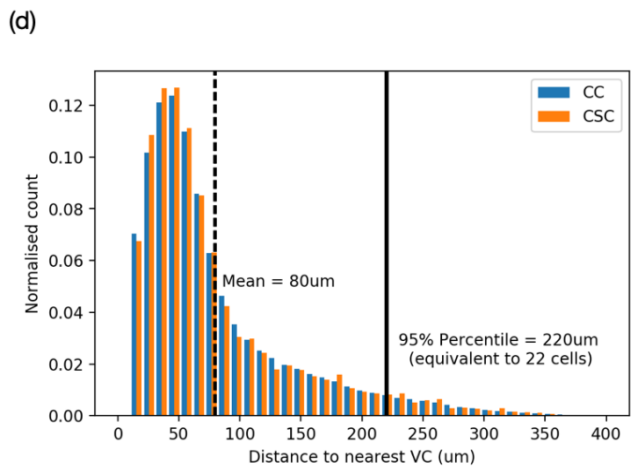

(e)

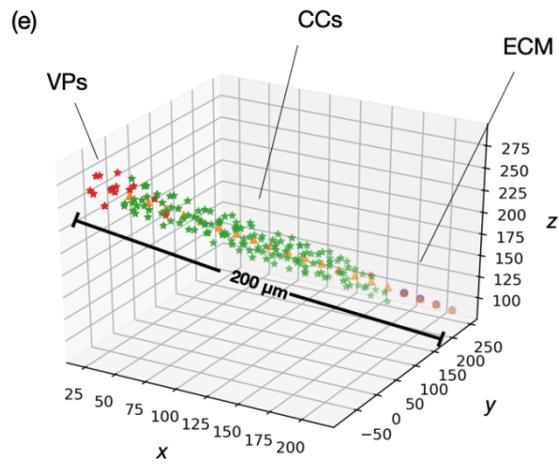

(f)

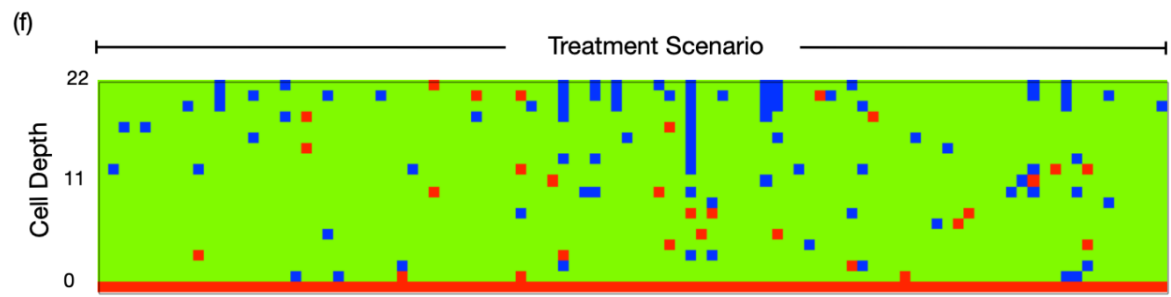

**Figure 2:** Virtual tumour module of the EVONANO platform. Tumours are grown *in silico* by (a) initialising tumour with several cells and vessel points and running simulation until the tumour population is sufficiently large. All simulations are performed using PhysiCell, where (b-c) show example outputs of a tumour with VPs and oxygen gradients where VPs are represented by red circles, green circles are CCs and brown circles are CC undergoing necrosis due to oxygen depletion. We use simulation output to (d) calculate the distance required for nanoparticles to penetrate from VPs and cover 95% of the tumour as well as (e) construct treatment scenarios such as those shown in (f), where red squares represent VPs, green squares represent CCs and blue squares represent ECM.

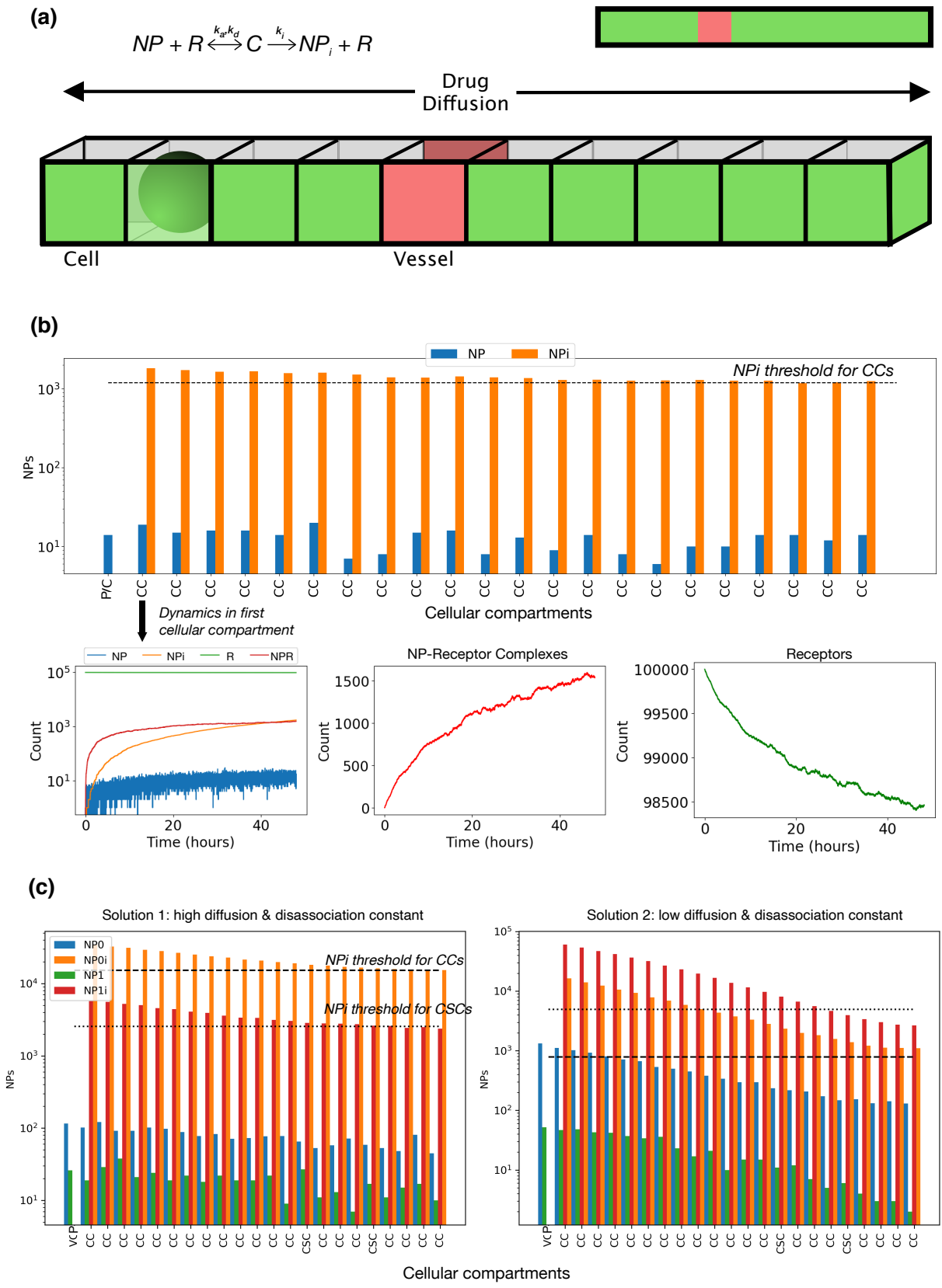

**Figure 3.:** Virtual Tissue Module of the EVONANO platform: We use STEPS to solve the stochastic reaction network, shown in (a), for nanoparticles that diffuse from vessel points into well mixed compartments representing tumour cells. In (b) we show the total penetration profile of nanoparticles within a homogenous tumour, such as the number of nanoparticles (NP) and internalised nanoparticles (NPi) within each cellular compartment. Here the nanoparticle and treatment parameters are able to successfully penetrate and kill all cancer cells. Each cell contains example Cell-nanoparticle dynamics such as increased number of nanoparticle-receptor complexes (NPR), and a decrease in overall receptors over time. We also show, in (c) the two optimum solutions of a heterogeneous cell population, as described in Section 4.

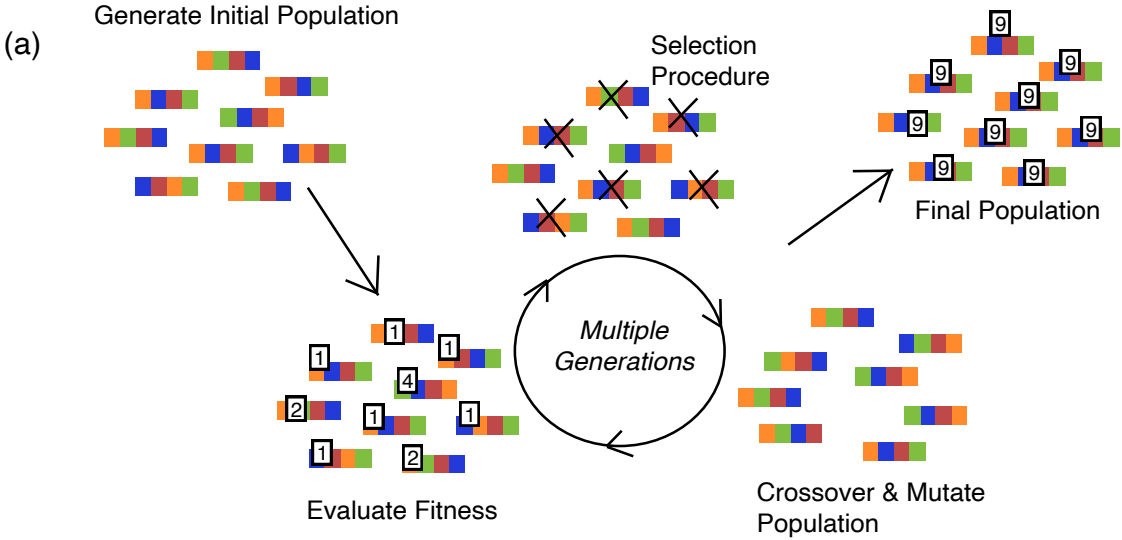

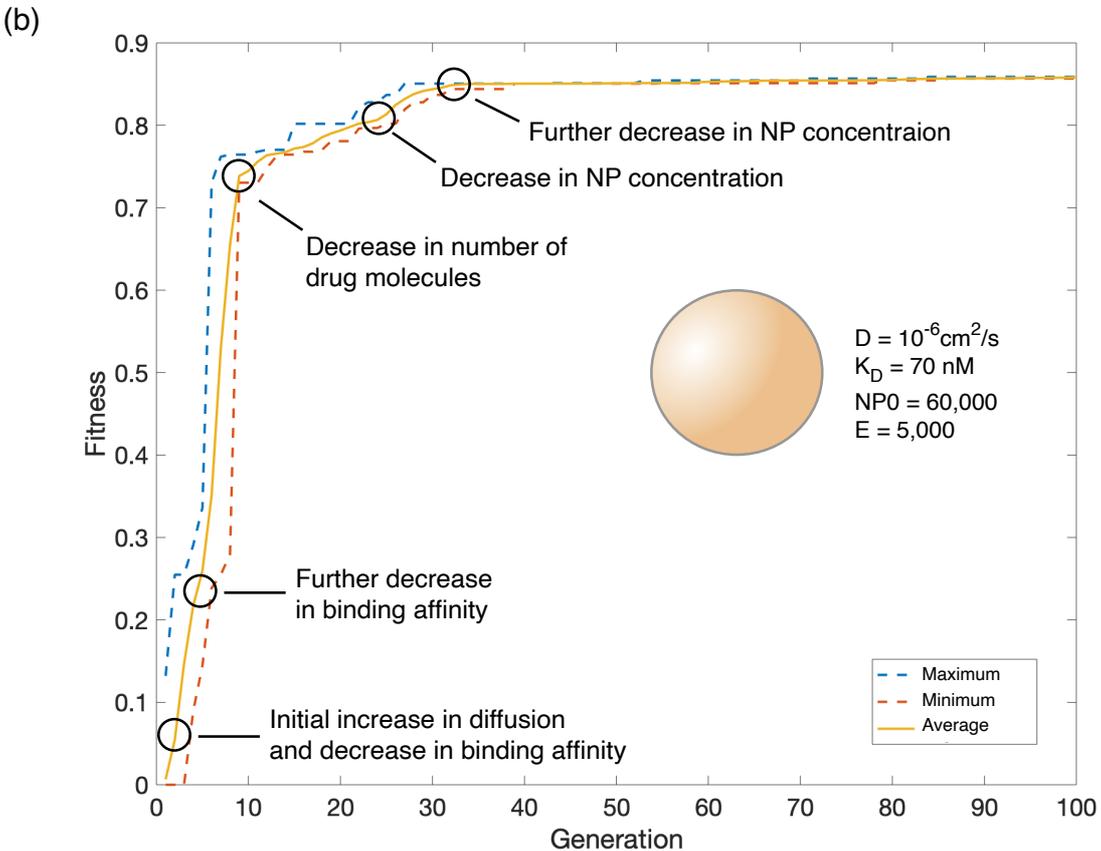
29

**Figure 4:** Machine Learning Module of the EVONANO platform: In this paper, we use (a) an evolutionary algorithm to optimise nanoparticle design. This optimisation routine first initialises a population of different parameter sets and evaluates the fitness of each parameter sets using the tissue module. Parameter sets with high fitness are selected for (using a tournament procedure) and these sets are used to generate a new population of nanoparticle parameters using the cross-over operator (where the information of the two parent solutions are combined to make new solutions), and mutated. This process is repeated for a prescribed number of generations, resulting in an overall increase in the fitness of the parameters and effective nanoparticle treatments. We show in (b) the mean, maximum and minimum change in fitness across generations and highlight the phenotypical changes of importance in parameters as we optimise the nanoparticle treatment.